\documentclass[
]{ceurart}

\usepackage{listings}
\usepackage{inputenc}
\begin{document}

\copyrightyear{2022}
\copyrightclause{Copyright for this paper by its authors.
  Use permitted under Creative Commons License Attribution 4.0
  International (CC BY 4.0).}

\conference{MediaEval'22: Multimedia Evaluation Workshop,
  January 13--15, 2023, Bergen, Norway and Online}

\title{Floods Relevancy and Identification of Location from Twitter Posts using NLP Techniques}


\author[1]{Muhammad Suleman} \fnmark[1] 

\author[1]{Muhammad Asif} \fnmark[1] 

\author[2]{Tayyab Zamir}\fnmark[1]
\author[1]{Ayaz Mehmood} \fnmark[1]
\author[3]{Jebran Khan}
\author[1]{Nasir Ahmad}
\author[4]{Kashif Ahmad} [%
    email=kashif.ahmad@mtu.ie
]

\address[1]{DCSE, University of Engineering and Technology, Peshawar, Pakistan}
\address[2]{Abasyn University Islamabad Campus, Pakistan}
\address[3]{Department of AI, AJOU University, South Korea}
\address[4]{Department of Computer Science, Munster Technological University, Cork, Ireland}

\cortext[1]{Corresponding author.}
\fntext[1]{These authors contributed equally.}

\begin{abstract}
This paper presents our solutions for the MediaEval 2022 task on DisasterMM. The task is composed of two subtasks, namely (i) Relevance Classification of Twitter Posts (RCTP), and (ii) Location Extraction from Twitter Texts (LETT). The RCTP subtask aims at differentiating flood-related and non-relevant social posts while LETT is a Named Entity Recognition (NER) task and aims at the extraction of location information from the text. For RCTP, we proposed four different solutions based on BERT, RoBERTa, Distil BERT, and ALBERT obtaining an F1-score of 0.7934,
0.7970, 0.7613, and 0.7924, respectively. For LETT, we used three models namely BERT, RoBERTa, and Distil BERTA obtaining an F1-score of 0.6256, 0.6744, and 0.6723, respectively.
 
\end{abstract}

\maketitle

\section{Introduction}
\label{sec:intro}
Natural disasters represent hazardous events that are generally caused by geophysical, hydrological, climatological, and meteorological elements. These hazardous events may have an adverse impact on human lives and infrastructure. Floods are one such event and it frequently occurs in different parts of the world. Similar to other natural disasters, floods may have a significant impact on public health and infrastructure. For instance,  it has been noticed on numerous occasions that roads and communication infrastructure are badly damaged during floods \cite{nguyen2017damage}. 

A rapid and effective response to disasters may help in mitigating their adverse impact. Access to relevant and timely information is critical for an effective response. The literature demonstrates several situations where access to relevant information may be possible due to several reasons, such as the unavailability of reporters in the area and damage to communication \cite{said2019natural}. Recently social media and crowdsourcing have been explored as a source of communication, information collection, and dissemination in emergency situations. To this aim, several interesting solutions have been proposed to collect, analyze, and extract meaningful insights from social media content \cite{said2019natural}. However, social media content also comes with several limitations. For instance, social media content is generally noisy, thus, making access to relevant information very challenging. Similarly, geolocation information, which is critical for the relevance of the content, is not necessarily available for all the relevant posts.  

Considering the importance and applications of social media content in disaster analytics floods detection in social media content has been also included in the MediaEval benchmark competition as a shared task for several years. This paper presents a solution for the MMDisaster task presented in MediaEval 2022 \cite{andreadis2022disastermm}. The challenge aims to solve two key challenges to disaster analytics in social media. The first subtask aims at reducing social media noise by automatically filtering social media content to obtain relevant content. The second subtask aims at extracting location information from social media text, allowing automatic positioning of a potential incident due to floods. For both subtasks, we proposed several interesting solutions as described in Section \ref{sec:approach}.

\section{Related Work}
\label{sec:work}
In recent years, the potential of social media has been widely explored in different application domains \cite{ahmad2019social,alsmadi2022adversarial}. Some of the key applications where social media content has been already proven very effective include public health \cite{andreadis2021social}, education \cite{al2021influence}, and public resource management \cite{ahmad2022social}. Social media outlets have also been widely explored for a diversified set of applications in disasters and emergency situations \cite{said2019natural}. For instance, Hao et al. \cite{hao2020leveraging} proposed a multi-modal framework utilizing multi-social media imagery and textual information for damage assessment in disaster-hit areas. The key factors analyzed in the work include hazard/disaster type, severity, and damage type. Wu et al. \cite{wu2018disaster} also utilized social media data and the associated geo-location information for generating early warnings and damage assessment analysis after disasters. Ahmad et al. \cite{ahmad2019automatic}, on the other hand, used social media imagery for the analysis of road conditions after the floods. More specifically, the authors proposed an early and late-fusion framework to identify passable roads in flooded regions. Alam et al. \cite{alam2018crisismmd} explored the potential of social media content in another relevant task of assessing flood severity. To this aim, the authors collected a large-scale benchmark dataset namely CrisisMD. The dataset provides a large collection of Twitter posts including textual and visual content. Hassan et al. \cite{hassan2022visual} explored a slightly different aspect of natural disasters by extracting sentiments and emotions from visual content shared in social media outlets. The authors detailed how visual sentiment analysis of disaster-related social media visual content can be utilized by different stakeholders, such as news agencies, public authorities, and humanitarian organizations.

Despite being proven very effective in different tasks of disaster analytics, social media content has several limitations, such as noisy data and the unavailability of geolocation information. In this paper, we propose a solution to overcome such challenges.
\section{Approach}
\label{sec:approach}

\subsection{Relevance Classification of Twitter Posts (RCTP)}
As a first step, we analyzed the available multimedia content. During the analysis, we observed that most of the posts missing visual content. Moreover, most of the images were irrelevant. Thus, we decided to use textual information only in our solution. Our framework for the RCTP subtask is composed of two steps. In the first step, we performed some pre-processing techniques to clean the data by removing unnecessary information, such as usernames, URLs, emojis, and stop words. 


After pre-processing, several state-of-the-art NLP algorithms including BERT \cite{devlin2018bert}, Roberta \cite{liu2019roberta}, Distil BERT \cite{sanh2019distilbert}, and ALBERT \cite{lan2019albert} are used for the classification of the text. Since its a binary classification task, in all methods, our cost function is based on binary crossentropy. Moreover, we used Adam optimizer with a mini batch size of 32 for 20 epochs. 

\subsection{Location Extraction from Twitter Texts (LETT)}
LETT subtask is treated as Named Entity Recognition (NER) task. NER involves locating and classifying named entities in text into pre-defined categories \cite{li2020survey}. In this task, we are interested in the identification of words representing the starting and subsequent words of a text sequence referring to a location. In LETT, annotations are provided at the word level. Similar to the RCTP task, in this task, we rely on multiple state-of-the-art algorithms including BERT, Roberta, Distil BERT, and ALBERT. We note that in this task, since annotations are provided at the word level, we did not use any pre-processing technique before training our models.

\subsection{Dataset}
For both subtasks, separate datasets are released. The dataset for RCTP subtask contains data from a total of 8,000 tweets. The tweets are collected between May 25, 2020, and June 12, 2020, using flood-related keywords in the Italian Language, such as ''alluvione'',  ''allagamento'', and ''esondazione''. 
The dataset is provided in two different sets namely the development set and the test set. The development set is composed of 5,337 tweets while the test set contains a total of 1,315 tweets.  

The dataset for the LETT subtask is composed of around 6,000 tweets collected between March 25, 2017, and August 1, 2018, using flood-related  Italian keywords. The annotations for this subtask are available per word in the tweets. 

\section{Results and Analysis}
\label{sec:results}

\subsection{Runs Description of RCTP Subtask}
Table \ref{tab:RCTP_Dev} shows the experimental results of the proposed solutions on the development set. We note that during the experiments on the development set, we used 70\% samples of the development set for training, 20\% for testing, and 10\% samples for validation. As can be seen in the table, no significant differences can be observed in the performance of the models on the clean and un-clean datasets. As far as the performance of the individual models is concerned, slightly better results are obtained with BERT compared to the other models.
\begin{table}[]
\caption{Experimental results of RCTP task on the development set.} 
\label{tab:RCTP_Dev}
\begin{tabular}{ccc}
\toprule
\textbf{Model} & \textbf{F1-Score on the Clean Dataset} & \textbf{F1-Score on the Un-clean Dataset} \\ 
\midrule
BERT    &  0.95 & 0.94 \\ 
RoBERTa  &   0.94 & 0.93 \\ 
Distil BERT  &   0.93 & 0.93 \\ 
ALBERT &   0.92 & 0.92 \\ 
 \bottomrule
\end{tabular}
\end{table}
Table \ref{tab:RCTP_test} provides the official results of the proposed solutions on the test set. We note that for the experiments on the test set the models are trained on the complete development set. In total, 4 different runs are submitted for the task. Our first, second, and fourth runs are based on BERT, RoBERTa, and Distil Bert models trained on the un-cleaned dataset, respectively. Our third run is based on the BERT model trained on the cleaned dataset. The performance of the models trained on the un-cleaned dataset is higher than the models trained on the cleaned dataset. This indicates that the pre-processing information resulted in the removal of some relevant features and thus has a negative impact on the results.

\begin{table}[]
\caption{Experimental results of the RCTP task on the test set.} 
\label{tab:RCTP_test}
\scalebox{.7}{
\begin{tabular}{cccc}
\toprule
\textbf{Run} & \textbf{Precision} & \textbf{Recall} & \textbf{F1-Score} \\ 
\midrule
 1 (BERT on Un-clean Dataset)& 0.6949 & 0.9251 & 0.7934 \\ 
 2 (RoBERTa on Un-lean Dataset) & 0.6947 & 0.9347 & 0.7970 \\ 
 3 (BERT on Clean Dataset) & 0.6486 & 0.9213 & 0.7613 \\ 
 4 (Distil BERT on Un-clean Dataset) & 0.6940 & 0.9232 & 0.7924 \\ 
 \bottomrule
\end{tabular}}
\end{table}
\subsection{Runs Description of LETT Subtask}
Table \ref{tab:LETT_Dev} provides the experimental results of the proposed solutions on the development set for the LETT subtask. Similar to RCTP, we used 70\% samples of the development set for training, 20\% for testing, and 10\% samples for validation. A significant variation can be observed in the results of the models on the development set. Overall, better results are obtained for RoBerta with a significant improvement of 3\% over the second-highest results obtained with the BERT model.
\begin{table}[]
\caption{Experimental results of the LETT task on the development set.} 
\label{tab:LETT_Dev}
\scalebox{.7}{
\begin{tabular}{cc}
\toprule
\textbf{Model} & \textbf{F1-Score} \\ 
\midrule
BERT    &  0.7752 \\ 
RoBERTa  &   0.8014 \\ 
Distil BERT  &   0.7658  \\ 
ALBERT &   0.6827  \\ 
 \bottomrule
\end{tabular}}
\end{table}
Table \ref{tab:LETT_test} provides the official results for the LETT subtask. In the experiments on the test set, the models are trained on the complete development set. We note that in the partial results the omitted samples are counted as false while in the partial results the omitted samples are completely ignored without any penalty. As can be seen in the table, overall, better results are obtained with Roberta and Distil BERT compared to the original implementation of the BERT model.

\begin{table}[]
\caption{Experimental results of LETT task on the test set.} 
\label{tab:LETT_test}
\scalebox{.7}{
\begin{tabular}{ccccccc}
\toprule
\multirow{2}{*}{\textbf{Run}} & \multicolumn{3}{c}{\textbf{Exact Results}} & \multicolumn{3}{c}{\textbf{Partial Results}} \\ \cline{2-7} 
 & \multicolumn{1}{c}{Precision} & \multicolumn{1}{c}{Recall} & F1-Score & \multicolumn{1}{c}{Precision} & \multicolumn{1}{c}{Recall} & F1-Score \\ 
 \midrule
 1 (BERT)& \multicolumn{1}{c}{0.596} & \multicolumn{1}{c}{0.522} & 0.556 & \multicolumn{1}{c}{0.628} & \multicolumn{1}{c}{0.622} & 0.625 \\ 
 2 (RoBERTa)& \multicolumn{1}{c}{0.540} & \multicolumn{1}{c}{0.676} & 0.600 & \multicolumn{1}{c}{0.577} & \multicolumn{1}{c}{0.810} & 0.674 \\ 
3 (Distil BERT) & \multicolumn{1}{c}{0.563} & \multicolumn{1}{c}{0.604} & 0.583 & \multicolumn{1}{c}{0.610} & \multicolumn{1}{c}{0.760} &  0.677\\ 
\bottomrule
\end{tabular}}
\end{table}
\section{Conclusions}
In this paper, we presented our solutions for the DisasterMM challenge posted in MediaEval 2022. For both subtasks, multiple state-of-the-art NLP algorithms are employed. In the current implementation, all the models are used individually, however, we believe these models can complement each other if jointly utilized in a merit-based fusion method. In the future, we aim to employ different merit-based fusion methods to jointly utilize the capabilities of the individual models in both subtasks.

\bibliography{references} 

\end{document}